\documentclass{article}
\usepackage{spconf,amsmath}
\usepackage{amsfonts}
\usepackage{csquotes}
\usepackage{multirow}
\usepackage{multicol}
\usepackage{siunitx}
\usepackage{textcomp}
\usepackage{tabularx}
\usepackage[pdftex]{graphicx}
\usepackage{caption}

\title{handling class imbalance in low-resource dialogue systems by combining few-shot classification and interpolation}

\name{Vishal Sunder and Eric Fosler-Lussier} %\thanks{Thanks to XYZ agency for funding.}
\address{The Ohio State University}
\begin{document}
%\ninept
%
\maketitle
\begin{abstract}
Utterance classification performance in low-resource dialogue systems is constrained by an inevitably high degree of data imbalance in class labels. We present a new end-to-end pairwise learning framework that is designed specifically to tackle this phenomenon by inducing a few-shot classification capability in the utterance representations and augmenting data through an interpolation of utterance representations. Our approach is a general purpose training methodology, agnostic to the neural architecture used for encoding utterances. We show significant improvements in macro-F1 score over standard cross-entropy training for three different neural architectures, demonstrating improvements on a Virtual Patient dialogue dataset as well as a low-resourced emulation of the Switchboard dialogue act classification dataset.
\end{abstract}
\section{Introduction}
\label{sec:intro}

In recent years, there has been a lot of interest in the deployment of question answering (QA) dialogue agents for specialized domains \cite{danforth2009development, danforth2013can, khurana2017hybrid}. A simple yet effective approach to this application has been to treat question answering as a  utterance classification task. Dialogue datasets annotated for this purpose typically have a large number of classes catering to very fine grained user queries. Due to the limited amount of data that can be realistically collected for specialized domains, the dataset becomes highly class-imbalanced, following a Zipfian-style distribution of utterance classes.

% \begin{figure}[htb]
%     \hfill
%     \centering
%     \centerline{\includegraphics[width=\columnwidth]{figures/model_c.eps}}
% \caption{Model overview. Sentences $x_i$ and $x_j$ are fed into the same encoder and interpolated thereafter. Solid lines indicate forward-propagation. Dashed lines indicate gradient flow.}
% \vspace*{-0.15in}
% \label{fig:model_over}
% \end{figure}

\begin{figure}[htb]
    \hfill
    \centering
    \centerline{\includegraphics[width=\columnwidth]{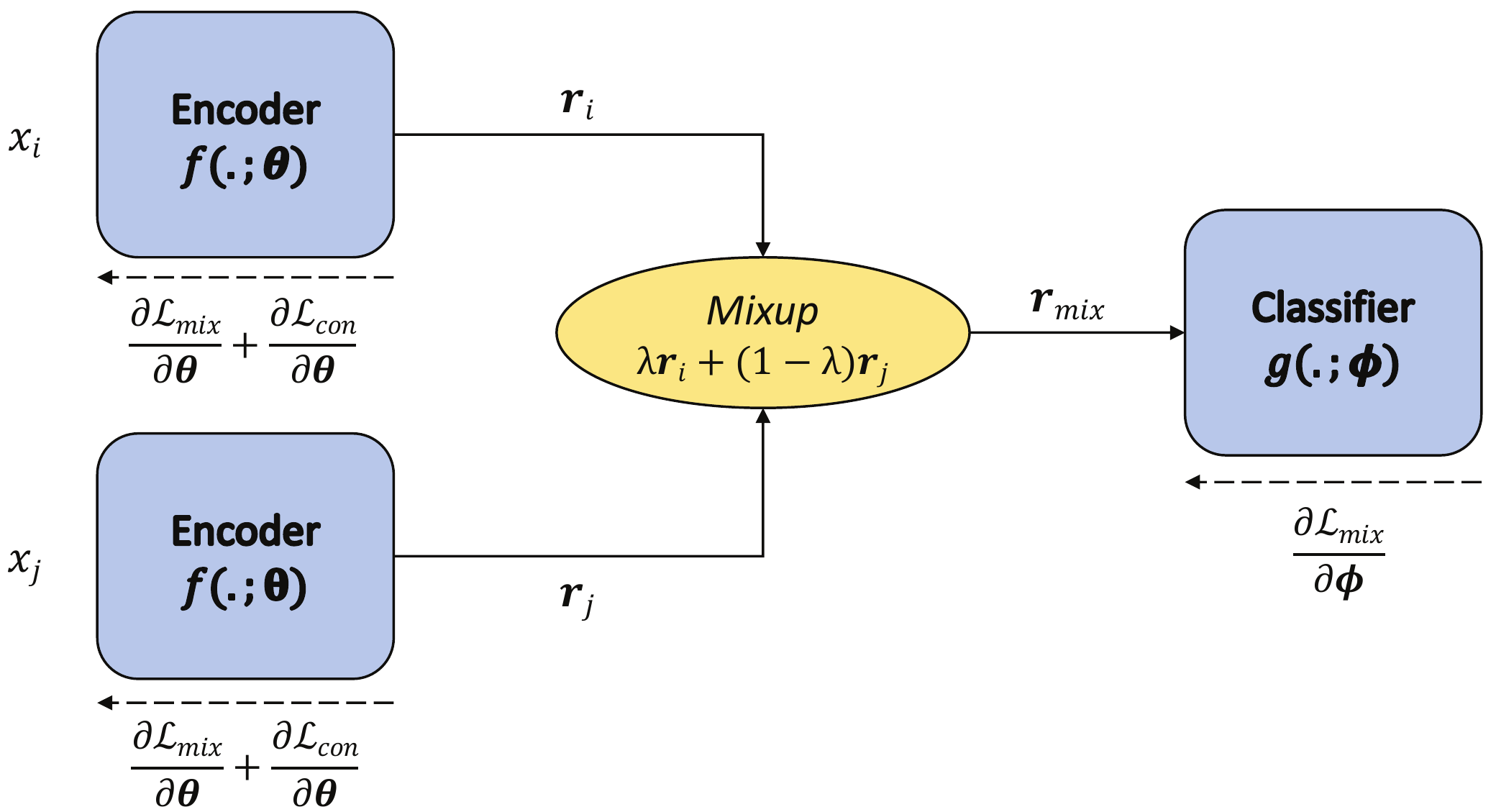}}
\caption{Model overview. Sentences $x_i$ and $x_j$ are fed into the same encoder and interpolated thereafter. Solid lines indicate forward-propagation. Dashed lines indicate gradient flow.}
\vspace*{-0.15in}
\label{fig:model_over}
\end{figure}

This challenge is very evident in the Virtual Patient dialogue agent \cite{danforth2009development, danforth2013can} used to train medical students at Ohio State to take patient histories. Student-posed questions are classified as one of 348 question types, ranging from the frequent ``How are you today?'' to the infrequently asked ``Are you nervous?''  The simple dialog strategy allows medical professors to author QA pairs for particular virtual patients without requiring them to become dialogue system experts.

Various methods have been proposed to handling rare classes in this low-resource dataset, including memory and paraphrasing \cite{jin2018using}, text-to-phonetic data-augmentation \cite{stiff2019improving} and an ensemble of rule-based and deep learning based models \cite{jin2017combining}. Recently, self-attention has shown to work particularly well for rare class classification \cite{stiff2020self}.

% Handling class-imbalance and learning from limited amount of data have been subjects of investigation in the field of machine learning in general. Pairwise training has shown promise in \textit{few-shot} classification tasks in computer vision. Recently, it has been shown to work well for text classification in combination with standard cross-entropy training. Ideas of interpolation from computer vision have been shown to work well in semi-supervised text classification tasks. However, these approaches assume the availability of unlabelled data which is hard to get for a specialized dialog system. 

% In this paper, we propose a novel end-to-end pairwise learning framework which combines the few-shot learning paradigm \cite{koch2015siamese} with an interpolation based data augmentation technique \cite{chen2020mixtext} to tackle the said problem of class-imbalance. Few-shot classification helps in identifying classes with as few as one example in the training set and we achieve this by using a contrast based training criterion with a nearest neighbor search. Further, pairs of training instances are interpolated using \textit{mixup} \cite{zhang2017mixup} for augmenting the data. This further helps to maintain performance on the frequent classes which has been lacking in other pairwise learning frameworks \cite{jin2018using} that end up needing additional model ensembling.

%In this paper, we 
We propose a novel end-to-end pairwise learning framework (Figure~\ref{fig:model_over}, described in Section~\ref{sec:pair_train}) which %combines 
augments
few-shot classification \cite{koch2015siamese} with an interpolation based data augmentation technique \cite{chen2020mixtext} to tackle the said problem of class-imbalance.\footnote{Code and data available at https://github.com/OSU-slatelab/vp-pairwise} Few-shot classification helps in identifying classes with as few as one example in the training set with a nearest neighbor search. Further, pairs of training instances are interpolated using \textit{mixup} \cite{zhang2017mixup} for augmenting the data. A classifier trained %on this data 
with augmented data
helps to maintain performance of the model on the frequent classes,
%. This has been lacking in 
unlike other pairwise learning frameworks \cite{jin2018using} that 
%end up needing 
require
additional model ensembling to maintain overall performance.

% The few-shot learning is achieved by combining a \textit{contrastive loss} with a nearest neighbor search which helps to classify the cases with as few as one example in the training set. Further, pairs of training instances are interpolated using \textit{mixup} \cite{zhang2017mixup} which augments the scarce data while helping to maintain performance on the frequent classes. Not being able to maintain performance on frequent classes has been a major drawback of other pairwise learning frameworks \cite{jin2018using} which end up requiring additional ensembling. We also perform ablation studies to show the effectiveness of each of these components.

% This type of data augmentation has shown promise in semi-supervised text classification \cite{chen2020mixtext}. But unlike previous work, we do not have access to any additional unlabelled data for training. A major setback about previously proposed pairwise learning frameworks \cite{jin2018using} is that the model is unable to retain its performance on frequent classes and hence ends up requiring additional ensembling. Our end-to-end framework not only performs better on rare classes but also successfully retains performance on frequent ones. 

%We show the effectiveness of this method in dealing with class imbalance in the virtual patient dialogue domain across a range of encoding techniques. To showcase the applicability of our approach to other similar dialogue domains, we also test it on a low-resource version of the Switchboard dialogue act classification dataset that has a similar class imbalance; 
The effectiveness of this method is demonstrated both in the virtual patient dialogue domain as well as a low-resource version of the Switchboard dialogue act classification task that has a similar class imbalance.
Our training approach considerably improves performance of three neural encoding systems over standard cross-entropy training on both datasets.

\section{Pairwise Learning Framework}
\label{sec:pair_train}

Our pairwise learning framework seeks to create representations of sentence pairs, $\textbf{r}_i$ and $\textbf{r}_j$, that are closer together when the sentences are of the same class, and further apart when of different classes, while still retaining good classification ability.
From an original training set of (sentence,class) pairs $\{(x_i, y_i)\}_{i=1}^{N}$, we sample 
a paired training set $\{(x_{i,1}, x_{i,2}, y_{i,1}, y_{i,2})\}_{i=1}^{N_s}$ using a strategy explained in Section \ref{subsec:sampling}.
As illustrated in Figure~\ref{fig:model_over}, our model uses one of three previously-developed encoders $f(\cdot;\boldsymbol\theta)$ to transform sentences $x_i$ into representations $\textbf{r}_i$; pairs of representations are then augmented through a {\em mixup} strategy that tries to disentangle classes during classification.

\subsection{Encoder}
\label{subsec:encoder}
The encoder is a deep neural network, $f(\cdot;\boldsymbol\theta)$, which takes the input $x_i$ and returns its vector representation $\textbf{r}_i = f(x_i;\boldsymbol\theta)$.
We apply this transformation to both utterances in an instance of the paired training data to obtain $\textbf{r}_i$ and $\textbf{r}_j$.
To model $f(\cdot;\boldsymbol\theta)$ we use three different neural architectures: Text-CNN, Self-attentive RNN and BERT. 
The first two encoders can run in realtime for  concurrent users; the last demonstrates a relatively recent state-of-the-art encoding technique.
\\

%\subsubsection{Text-CNN encoder \cite{jin2017combining, kim2014convolutional}}
%\label{subsubsec:cnn}
%We use 
\noindent \textbf{Text-CNN encoder \cite{jin2017combining, kim2014convolutional}:}
The Text-CNN encoder utilizes
convolutions on 300 dimensional GloVe embeddings \cite{pennington2014glove} of $x_i$ with 300 filters of size 1,2 and 3. 
Each filter's output is max-pooled over time; the concatenated poolings are
%The max-pooled over time output of each filter is concatenated and 
fed into a fully connected layer with $tanh$ activation to obtain the final encoded representation $\textbf{r}_i \in \mathbb{R}^{600}$. This operation is performed for both $x_i$ and $x_j$ in a paired training instance using the same CNN to obtain $\textbf{r}_i$ and $\textbf{r}_j$.
\\

%\subsubsection{Self-attentive RNN encoder %\cite{stiff2020self,lin2017structured}}
%\label{subsubsec:rnn}

\noindent \textbf{Self-attentive RNN encoder \cite{stiff2020self,lin2017structured}:}
The Self-attentive RNN is a bidirectional Gated Recurrent Unit (GRU) \cite{cho2014learning} with self attention. The embedding representations of $x_i$ and $x_j$ are passed separately through a self-attentive bidirectional-GRU to obtain the $n$-head representation matrices, $\textbf{M}_i,\textbf{M}_j  \in \mathbb{R}^{600\times n}$. Each column $\textbf{m}_{ik}$ (and $\textbf{m}_{jk}$) of $\textbf{M}_i$ (and $\textbf{M}_j$) is an attention head representation.

To give more importance to the attention heads with similar representations in a paired instance, we perform a novel \textit{$2^{nd}$ order attention} on the attention head representations to obtain final representations $\textbf{r}_i$ and $\textbf{r}_j$. Mathematically,

\begin{equation*}
\resizebox{.45\columnwidth}{!}{%
$\begin{aligned}
&score_k = tanh(\textbf{m}_{ik}^T \textbf{W} \textbf{m}_{jk}) \\
    &\textbf{s}_k = \dfrac{exp(score_k)}{\sum_{i=1}^n exp(score_i)} \\
    &\textbf{r}_i = \textbf{M}_i\textbf{s} \\
    &\textbf{r}_j = \textbf{M}_j\textbf{s}
\end{aligned}$
}
\enspace
\vline
\enspace
\vcenter{\hbox{\begin{minipage}{.48\linewidth}
\centering
\includegraphics[width=4cm]{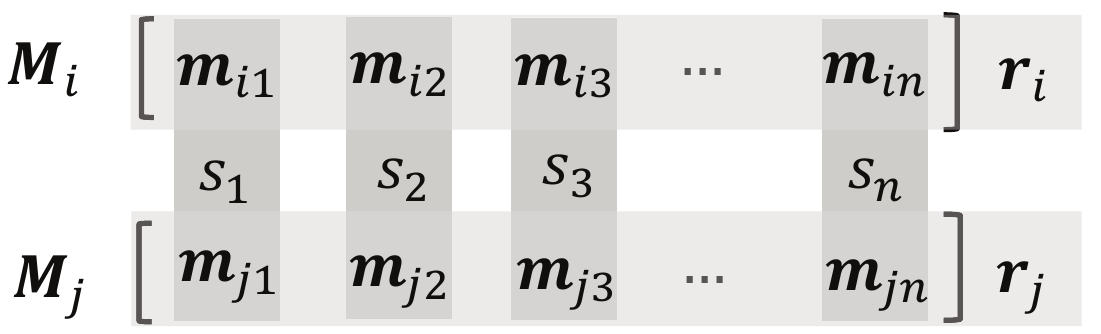}
\captionof{figure}{\textit{$2^{nd}$ order attention} highlights attention-head similarites in matched pairs.}
\label{fig:attention}
\end{minipage}}}
\end{equation*}

% \begin{equation*}
%     \begin{split}
%         &score_k = tanh(\textbf{m}_{ik}^T \textbf{W} \textbf{m}_{jk}) \\
%         &\textbf{s}_k = \dfrac{exp(score_k)}{\sum_{i=1}^n exp(score_i)} \\
%         &\textbf{r}_i = \textbf{M}_i\textbf{s} \\
%         &\textbf{r}_j = \textbf{M}_j\textbf{s}
%     \end{split}
% \end{equation*}

\noindent with $\textbf{W} \in \mathbb{R}^{600\times 600}$ a matrix of parameters learned during the training process; $\textbf{s} \in \mathbb{R}^n$ is a probability distribution that gives the weight of each attention-head representation. 
\\

%\subsubsection{BERT encoder \cite{devlin2018bert}}
%\label{subsubsec:bert}
\noindent\textbf{BERT encoder \cite{devlin2018bert}:}
We fine-tune the pretrained BERT model \textit{bert-based-uncased} 
%using the bert-based-uncased tokenizer and the 
using default WordPiece embeddings.\footnote{https://huggingface.co/transformers/model\textunderscore doc/bert.html} The final layer output of this encoder is a matrix $\textbf{B}_i \in \mathbb{R}^{768\times l}$ where $l$ is the length of the sequence. 
The encoded representations $\textbf{r}_i$ and $\textbf{r}_j$ are found by mean pooling over the columns of $\textbf{B}_i$ and $\textbf{B}_j$ respectively. 
% We did not perform the \textit{$2^{nd}$ order attention} on the BERT attention heads because it proved to be a large overhead in terms of the time and space complexity during test time.

\subsection{Classifier}
\label{subsec:classifier}
The classifier ($g(.;\boldsymbol\phi)$) is a 3-layer fully-connected MLP with $tanh$ activations. 
We use a \textit{mixup} strategy \cite{zhang2017mixup} to combine the representations, $\textbf{r}_i$ and $\textbf{r}_j$ in a paired training instance and feed it to the classifier. The idea is to create new training instances of data by interpolating between two instances of training examples. Formally,
\begin{equation*}
    \begin{split}
        \textbf{r}_{mix} &= \lambda \textbf{r}_i + (1-\lambda) \textbf{r}_j \\
        \textbf{y}_{mix} &= \lambda \textbf{y}_i + (1-\lambda) \textbf{y}_j
    \end{split}
\end{equation*}
Here, $\textbf{y}_i$ and $\textbf{y}_j$ are one-hot representations of $y_i$ and $y_j$ respectively. $\lambda$ is sampled from %a beta distribution as, 
$\lambda \sim \text{Beta}(\alpha, \alpha)$.
%, $\alpha$ is a hyperparameter that controls the distribution of $\lambda$. 
Higher values of the hyperparameter $\alpha$ result in $\lambda$ being close to 0.5. For rare classes, we tune $\alpha \in \{10.0,20.0\}$ to generate more novel data instances around the rare class. To preserve the frequent class distribution, we tune $\alpha \in \{0.05,0.1\}$ for frequent classes generating $\lambda$ close to 1.0. We found that this strategy performed better than using a fixed $\alpha$ value.

The classifier's output representation  
%takes the mixed representation
$\textbf{c}^{class}_{mix} = g(\textbf{r}_{mix};\boldsymbol\phi) \in \mathbb{R}^{C}$ is the pre-softmax output of the classifer.

%$\textbf{r}_{mix}$ and outputs the pre-softmax representation, $\textbf{c}^{class}_{mix} = g(\textbf{r}_{mix};\boldsymbol\phi) \in \mathbb{R}^{C}$.
%
%with $\textbf{c}^{class}_{mix} \in \mathbb{R}^{C}$ are the confidence scores over a set of $C$ classes.
% and $\textbf{p}^{class}_{mix} \in \mathbb{R}^{C}$ is the probability distribution over the same.

\subsection{Pairwise Loss Function}
\label{subsec:optim}
\subsubsection{Contrastive Loss}
\label{subsubsec:contrast}
To learn rich utterance representations for few shot classification, we use the contrastive loss \cite{hadsell2006dimensionality} on encoder outputs which helps to separate classes in the semantic space more effectively. For
%
% Given 
a paired training instance $(x_i,x_j,y_i,y_j)$ with 
$D$ being the euclidean distance between the normalized encoded representations of $x_i$ ($\frac{\textbf{r}_i}{\left\Vert \textbf{r}_i \right\Vert_2}$) and $x_j$ ($\frac{\textbf{r}_j}{\left\Vert \textbf{r}_j \right\Vert_2}$):
%, the contrastive loss \cite{hadsell2006dimensionality} is defined as,
\begin{equation*}
    \begin{split}
        \mathcal{L}_{con}(\boldsymbol\theta) &= \frac{1}{2}y_{pair}\{max(0,D-m_{pos})\}^2 \\&+ \frac{1}{2}(1-y_{pair})\{max(0,m_{neg}-D)\}^2
    \end{split}
\end{equation*}
where $y_{pair}$ is set to $1$ if $y_i = y_j$ and $0$ otherwise. %$D$ is the euclidean distance between the normalized encoded representations of $x_i$ ($\frac{\textbf{r}_i}{\left\Vert \textbf{r}_i \right\Vert_2}$) and $x_j$ ($\frac{\textbf{r}_j}{\left\Vert \textbf{r}_j \right\Vert_2}$). 
$m_{pos}$ and $m_{neg}$ are the positive and negative margins respectively. We use $m_{pos}=0.8$ and $m_{neg}=1.2$.

\subsubsection{Mixup Loss}
\label{subsubsec:mixup}
We train the classifier to predict the mixed class representation $\textbf{y}_{mix}$ (defined in section \ref{subsec:classifier}) by using the KL-divergence between $\textbf{y}_{mix}$ and the classifier predicted distribution as the second component of the loss function i.e.,
\begin{equation*}
    \begin{split}
        \mathcal{L}_{mix}(\boldsymbol\theta,\boldsymbol\phi) = \text{KL}(\textbf{y}_{mix} \Vert softmax(\textbf{c}^{class}_{mix}))
    \end{split}
\end{equation*}

\noindent We found that KL-divergence works better than cross-entropy on mixed labels for the datasets we used.
We combine the two losses using a hyperparameter $\beta\in[0,1]$:
%The final loss function for pairwise training is then defined as
\begin{equation*}
    \begin{split}
        \mathcal{L}_{pair}(\boldsymbol\theta,\boldsymbol\phi) = \beta \mathcal{L}_{con}(\boldsymbol\theta) + (1-\beta) \mathcal{L}_{mix}(\boldsymbol\theta,\boldsymbol\phi)
    \end{split}
\end{equation*}
%Here, $\beta \in [0,1]$ is a hyperparameter. 

\subsection{Testing}
\label{subsec:testing}

At test time, an utternace $x_{test}$ is encoded to obtain $\textbf{r}_{test} = f(x_{test};\boldsymbol\theta)$.\footnote{For the self-attentive RNN, we just average the attention head representations during test time as we don't have a paired counterpart to perform the \textit{$2^{nd}$ order attention}.} %
For each class, we perform a 1-nearest-neighbor search\footnote{For efficient search, we utilize the FAISS toolkit \cite{JDH17} (https://github.com/facebookresearch/faiss)} on the training set using $\textbf{r}_{test}$ and set the corresponding elements of the class score $\textbf{c}^{nn}_{test} \in \mathbb{R}^C$ to be the inverse distance to $\textbf{r}_{test}$. We also compute the classifier class scores on the unmixed test utterance, $\textbf{c}^{class}_{test} = g(\textbf{r}_{test};\boldsymbol\phi)$. 

Each of these confidence scores have distinct advantages for rare class classification. $\textbf{c}^{nn}_{test}$ incorporates a \textit{few-shot} classification capability on the rarest classes \cite{koch2015siamese} and $\textbf{c}^{class}_{test}$ incorporates a capability of using the classifier trained using the augmented data \cite{zhang2017mixup}.
We combine the two by first normalizing them and then interpolating:
\begin{equation*}
    \begin{split}
        &\textbf{c}^{final}_{test} = \gamma\bar{\textbf{c}}^{nn}_{test} + (1-\gamma)\bar{\textbf{c}}^{class}_{test} \\
    \end{split}
\end{equation*}
$\gamma \in [0,1]$ is tuned on the validation data. The maximal element of $\textbf{c}^{final}_{test}$ %is the final confidence which 
is used to make the prediction.

\section{Experimental Setup}
\label{sec:experimental_setup}
\subsection{Sampling of pairs}
\label{subsec:sampling}
%First, we 
We randomly sample 50,000 positive pairs ($y_i = y_j$) and 100,000 negative pairs ($y_i \neq y_j$) to create sets $\mathcal{S}^{+}$ and $\mathcal{S}^{-}$ respectively.
Once per epoch,
%Then, at the beginning of each epoch, 
for every pair $(x_i^+, x_j^+) \in \mathcal{S}^{+}$ we compute $d_{ij}^+ = \left \Vert f(x_i^+; \boldsymbol\theta) - f(x_j^+; \boldsymbol\theta) \right \Vert_2$ and select the top 25,000 pairs with the \textit{highest} corresponding $d_{ij}^+$. Similarly, we compute $d_{ij}^-$ for pairs $(x_i^-, x_j^-) \in \mathcal{S}^{-}$ and select the top 50,000 pairs with the \textit{lowest} corresponding $d_{ij}^-$. This gives us a paired training set size of 75,000.
%at the beginning of each epoch.

\subsection{Datasets}
\label{subsec:dataset}

\noindent \textbf{Virtual Patient Corpus   \cite{jin2017combining,stiff2020self}:}
%\subsubsection{Virtual Patient Corpus}
%\label{subsubsec:vp_corpus}
The Virtual Patient (VP) dataset is a collection of dialogues between medical students and a virtual patient experiencing back pain. 
%This is a spoken dialogue which is then manually transcribed. It was converted into a question answering dataset where 
Each student query  was mapped to a question from set of 348 canonical questions whose answers are already known.
The data consists of a total of 9,626 question utterances over 259 dialogues (6,827 training, 2,799 test). %There are a total of 348 canonical questions to which each of these questions are mapped. A total of 2,799 utterances are held out as the test set,
%. This leaves 
%leaving a total of 6,827 questions in the training set. 
%
The data are highly imbalanced across classes (Table~\ref{tab:quintile}), with the top fifth of classes comprising 70\% of the examples in the training set.
%This dataset is highly imbalanced. The top quintile of the 348 class labels has around 70\% examples in the training set. The entire distribution is shown in Table \ref{tab:quintile}.
\\

\noindent \textbf{Switchboard Dialog Act corpus \cite{jurafsky1997switchboard}:}
%\subsubsection{Switchboard Dialog Act corpus}
%\label{subsubsec:swda_corpus}
The Switchboard Dialog Act (SwDA) dataset  is a collection of telephone conversations which are transcribed and annotated for 43 dialog act classes;
%. To the best of our knowledge, it is the only utterance classification dataset with 
it exhibits a class imbalance similar to the Virtual Patient data (see Table \ref{tab:quintile}). 
SwDA has a training set size of 193k, validation set size of 23k and test set size of 5k. We experiment with several subsets to simulate data growth;
we also create a VP-style low-resource setting which has 5 subsets of 6850 instances each from the entire SwDA training set by random sampling. All the models are trained on these five data subsets and the mean \textpm\  standard deviation is reported.

\subsection{Training details}
\label{subsec:exp_details}
We train using 90/10 train/dev splits, using the Adam optimizer \cite{kingma2014adam} with a learning rate $4e-5$ for BERT and $1e-3$ for the rest. BERT is trained for 6 epochs; other models are trained for 15 epochs. The model-epoch with the best dev set Macro-F1 performance is retained for testing.

\section{Results and Analysis}
\label{sec:results_main}

\begin{table}
\resizebox{\columnwidth}{!}{\begin{tabular}{|c|ccccc|}
\hline
 Quintile \#  & $1^{st}$ & $2^{nd}$ & $3^{rd}$ & $4^{th}$ & $5^{th}$ \\
\hline
 
 Virtual Patient  & 1.8\% & 3.5\% & 6.9\% & 17.7\% & 70.1\%  \\ %\cline{2-10} 
 Switchboard  & 0.3\% & 0.9\% & 3.0\% & 6.6\% & 89.2\%  \\ %\cline{2-10} 
 \hline
\end{tabular}}
\caption{\% of data in the training set per class quintile}
\label{tab:quintile}
\vspace*{-0.15in}
\end{table}

\begin{table}[b]
\centering
\resizebox{\columnwidth}{!}{\begin{tabular}{|c|c|cc|cc|}
\cline{1-6}
 \multicolumn{1}{|c|}{\multirow{2}{*}{\begin{tabular}[c]{@{}c@{}}Encoder Model\end{tabular}}}   & \multicolumn{1}{|c|}{\multirow{2}{*}{\begin{tabular}[c]{@{}c@{}} Training \end{tabular}}}  & \multicolumn{2}{c|}{SwDA} & \multicolumn{2}{c|}{Virtual Patient} \\ \cline{3-6} 
    &   & Acc(\%)  & F1(\%)  & Acc(\%) & F1(\%)   \\ \hline
\multicolumn{1}{|c|}{\multirow{2}{*}{\begin{tabular}[c]{@{}c@{}}Text-CNN\end{tabular}}} & Cross-Entropy\cite{jin2017combining}   & 59.3 \textpm 0.7 & 26.4 \textpm 1.2 & 75.7 & 51.4 \\ %\cline{2-10} 
\multicolumn{1}{|c|}{} & $\mathcal{L}_{pair}$ & \textbf{60.8} \textpm 0.8 & \textbf{32.4} \textpm 0.9 & \textbf{75.8} & \textbf{57.1} \\ \hline
\multicolumn{1}{|c|}{\multirow{2}{*}{\begin{tabular}[c]{@{}c@{}}Self-attentive \\RNN\end{tabular}}} & Cross-Entropy\cite{stiff2020self} & 61.0 \textpm 1.2 & 29.4 \textpm 1.2 & \textbf{79.2} & 59.8 \\ %\cline{2-10} 
\multicolumn{1}{|c|}{} & $\mathcal{L}_{pair}$ & \textbf{61.9} \textpm 0.9 & \textbf{32.7} \textpm 1.0 & \textbf{79.2} & \textbf{63.9} \\ \hline
\multicolumn{1}{|c|}{\multirow{2}{*}{BERT}} & Cross-Entropy  & \textbf{65.2} \textpm 0.6 & 30.4 \textpm 1.7 & 79.4 & 57.4 \\ %\cline{2-10} 
\multicolumn{1}{|c|}{} & $\mathcal{L}_{pair}$ & 64.6 \textpm 0.5 & \textbf{35.5} \textpm 1.7 & \textbf{81.0} & \textbf{66.8} \\ \hline
\end{tabular}}
\caption{Comparing performance of $\mathcal{L}_{pair}$ training with cross-entropy training using 3 different models on the two datasets. For SwDA, we perform experiments with 5 smaller subsets of the training data. We report $mean \pm std$ of the performance on these subsets for SwDA. Bold represents the best performance.}
\label{tab:results_main}
\vspace*{-0.15in}
\end{table}

We compare the performance of the proposed pairwise training against the conventional cross-entropy training performance (Table~\ref{tab:results_main}).\footnote{In the virtual patient test set, a few classes are absent. Previous work \cite{stiff2020self} does not take this into account when computing the macro-F1 score and hence report a slightly underestimated value. We correct for this.} For each of the three neural architectures used for modeling the encoder, we train a corresponding model with the same architecture  using a cross-entropy loss. 
%The results\footnote{} are reported in 
%Table \ref{tab:results_main}.
Macro-F1 scores improve for all models on all datasets. 

%We see substantial improvements in terms of macro-F1 scores for all the models on all datasets. As the datasets are highly imbalanced, macro-F1 gives a better measure of which model has more successfully learned to distinguish rare classes from the frequent ones. 

\begin{figure}[tb]

\hfill

\begin{minipage}[b]{1.0\columnwidth}
  \centering
  \centerline{\includegraphics[width=1.1\columnwidth]{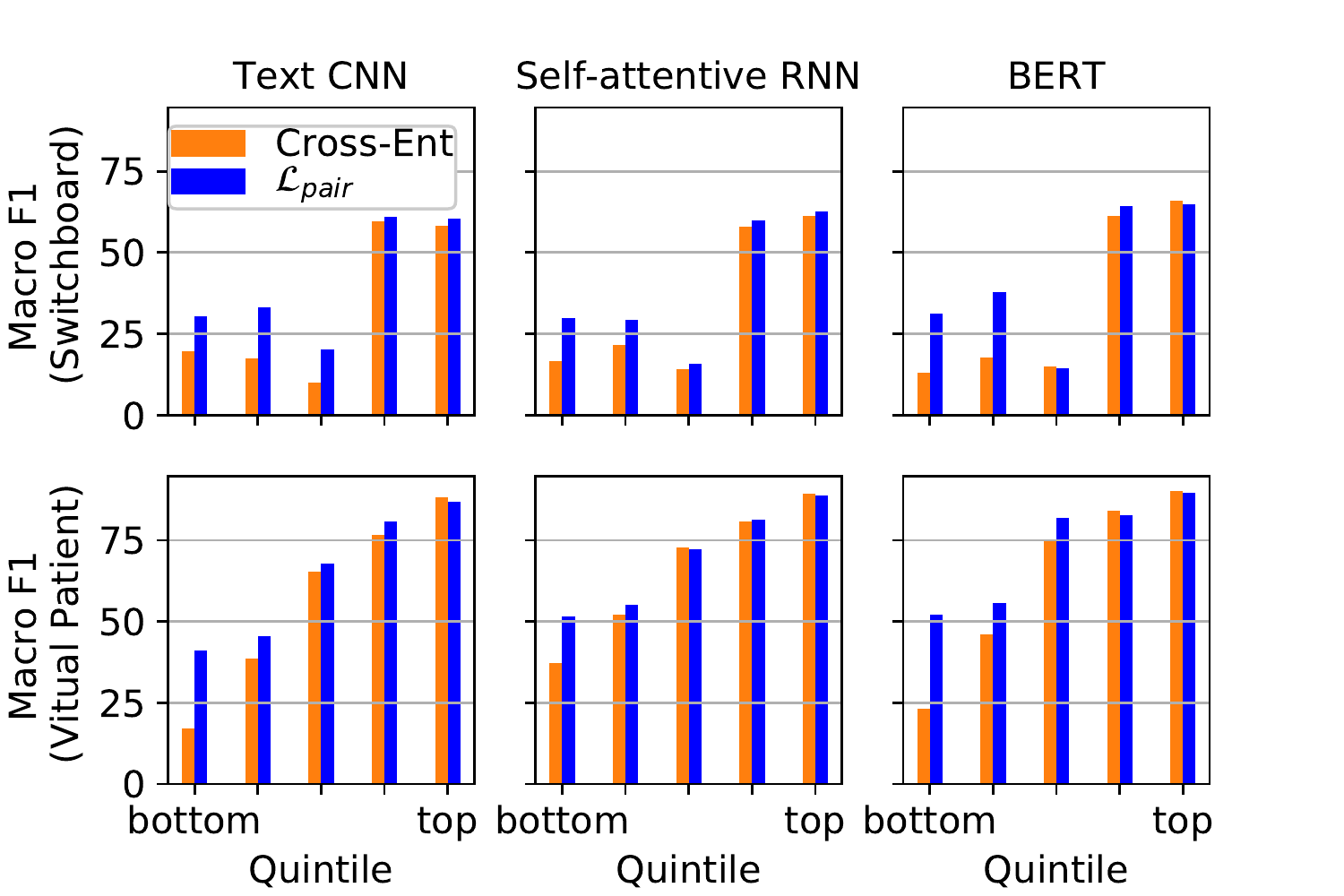}}
%  \vspace{2.0cm}
\end{minipage}

\caption{Quintile-wise performance of different models on the two datasets. Pairwise training (blue) usually helps rare classes over cross-entropy training (orange) and provides similar performance for frequent classes across decoders. }
\label{fig:quint_wise}
\vspace*{-0.15in}
\end{figure}

We also plot macro-F1 performance for each class quintile separately (Figure~\ref{fig:quint_wise}).
%To see how each model performs when classifying class labels from the different quintiles, we plot their macro-F1 performance on each class quintile separately as shown in Figure \ref{fig:quint_wise}. %^From the plots, it 
It is clear that pairwise training does extremely well compared to cross-entropy training in the lower quintiles, while retaining the performance in the top quintile. %An important point to note is that 
In particular, the BERT based pairwise training yields an improvement of almost 2x on the bottom quintile. This is especially useful given that a major drawback of BERT has been its poor performance on rare classes (\cite{stiff2020self}, \cite{mahabal2020text}).

\begin{figure}[tb]

\hfill

\begin{minipage}[b]{1.0\linewidth}
  \centering
  \centerline{\includegraphics[width=8.5cm]{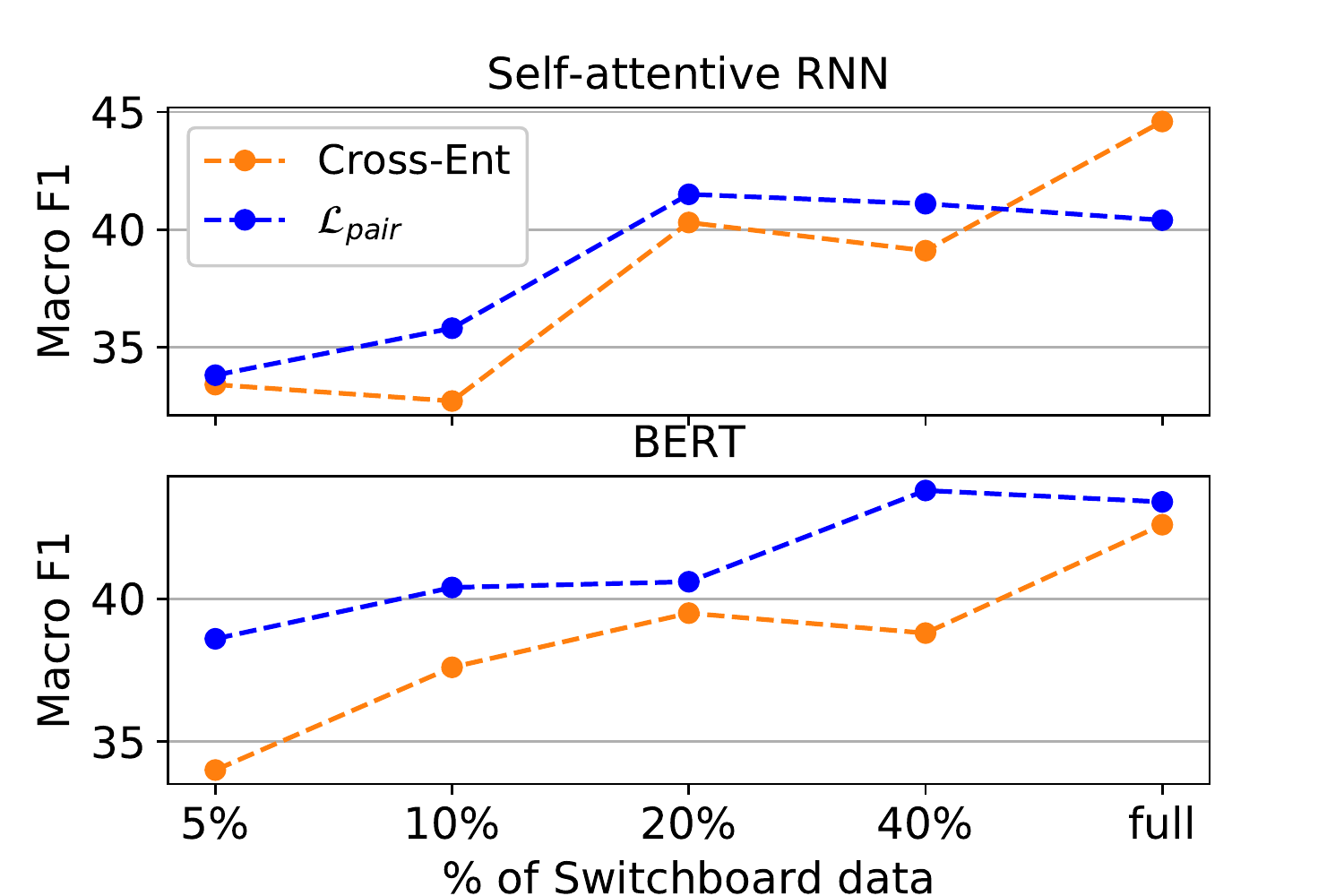}}
%  \vspace{2.0cm}
\end{minipage}

\caption{Effect on F1 of varying the amount of training data. We test using the self-attentive RNN and BERT as they were the best performing models for the Virtual Patient data.}
\label{fig:data_var}
\vspace*{-0.15in}
\end{figure}

We also compare the performance of the proposed pairwise training with cross-entropy training as we increase the training data for SwDA (Figure \ref{fig:data_var}). %We see that up
Up
through 40\% of the SwDA data (train set size of 77k), pairwise training gives better performance compared to cross-entropy training. With the full data (train set size of 193k), pairwise training does not perform quite as well for self-attentive RNN with minor improvements using BERT. %But the fact that 
That pairwise training does better for a data size as high as 77k utterances is encouraging, as task specific dialog datasets for real-world deployment will usually start at a much smaller size. 
% The Virtual Patient dataset is a very good example of this.

\begin{table}[b]
\vspace*{-0.1in}
\resizebox{\columnwidth}{!}{\begin{tabular}{|c|cc|cc|}
\cline{1-5}
 \multicolumn{1}{|c|}{}  & \multicolumn{2}{c|}{SwDA} & \multicolumn{2}{c|}{Virtual Patient} \\ \cline{2-5} 
 & Acc(\%)  & F1(\%) & Acc(\%) & F1(\%)  \\ \hline
 Full  & \textbf{61.9} \textpm 0.9 & \textbf{32.7} \textpm 1.0 & \textbf{79.2} & \textbf{63.9}  \\ %\cline{2-10} 
\hline
$-\mathcal{L}_{con} (\beta = 0; \gamma = 0)$ & 60.5 \textpm 1.0 & 32.1 \textpm 1.5 & 79.1 & 62.6 \\ 
$-\mathcal{L}_{mix} (\beta = 1; \gamma = 1)$ & 38.6 \textpm 0.7 & 22.9 \textpm 0.6 & 76.8 & 61.1 \\  \hline
$-$ \textit{$2^{nd}$ order attention} & \textbf{61.9} \textpm 0.8 & \textbf{32.7} \textpm 1.0 & 78.6 & 61.3 \\  \hline
\end{tabular}}
\caption{Ablation studies on self-attentive RNN on the two datasets, removing one either part of $\mathcal{L}_{pair}$. Results on SwDA are the $mean \pm std$ on 5 small subsets of the training data. %Bold represents best performance.
}
\label{tab:ablation}
\end{table}

Finally, we perform ablation studies to see the effectiveness of each component of $\mathcal{L}_{pair}$. We trained the self-attentive RNN by removing one component at a time (Table \ref{tab:ablation}). %We see that 
Both $\mathcal{L}_{con}$ and $\mathcal{L}_{mix}$ contribute to the final performance; $\mathcal{L}_{con}$ helps more on the VP dataset and not as much on SwDA. This may be because virtual patient has a lot more classes spanning across a similar training set size. Hence, many classes have as few as one instance in the training set. Therefore, using $\mathcal{L}_{con}$ with a 1-nearest-neighbor classification helps more in these few-shot cases. As our \textit{$2^{nd}$ order attention} method is new, we note through ablation that it helps with the VP data while making no difference on SwDA; we attribute this to two possible hypotheses: the classes in Virtual Patient are more semantically fine-grained in contrast to speech-act classes in SwDA, so attending to specific attention heads may be more crucial in Virtual Patient, or since SwDA utterances are relatively shorter, the information contained in different attention heads may be correlated.  This will be investigated more fully in subsequent work.

\section{Conclusion}
\label{sec:conclusion}
 We proposed an end-to-end pairwise learning framework which mitigate class imbalance issues in low-resource dialogue systems,
 by generalizing well to the rare classes while maintaining performance on the frequent ones. By using a combination of a contrast based and an interpolation based loss function, we show considerable improvements over %run-of-the-mill 
cross-entropy training. %for such dialogue systems. 
Effectively incorporating dialogue context in the proposed pairwise training is a subject of future work.

\section{Acknowledgements}
This material builds upon work supported by the National Science Foundation under Grant No. 1618336.  We gratefully acknowledge the OSU Virtual Patient team, especially Adam Stiff, for their assistance.
% Future work includes looking for effective ways to incorporate dialogue context in the proposed pairwise training.
% and also combining it with other data-augmentation techniques.

% We introduce a new pairwise learning framework to mitigate class imbalance issues in low-resource dialogue systems. Our training criterion is a combination of a contrast based and an interpolation based loss function. We show considerable improvements of this type of training over run-of-the-mill cross-entropy training in classifying rare classes. Future work includes looking for effective ways to incorporate dialogue context in pairwise training and also combining it with other approaches like paraphrasing and backtranslation.

% There is a lot of scope for extending this work particularly in the dialogue domain. An interesting direction in this domain could be looking for effective ways to incorporate dialogue context in pairwise training. Can we efficiently combine an interpolation based strategy with other data-augmentation approaches like paraphrasing and backtranslation is also an important question.

% Our methodology is agnostic to the underlying neural network model, hence any sophisticated neural architecture can be seamlessly integrated with our proposed framework.

\bibliographystyle{IEEEbib}
\bibliography{strings,refs}

\end{document}